%% file: iclr2020_conference.tex
\documentclass{article} 
\usepackage{iclr2020_conference,times}

\input{math_commands.tex}

\usepackage{hyperref}
\usepackage{url}
\usepackage{amsmath,graphicx}
\usepackage{commath}
\usepackage{ amssymb }
\usepackage{ mathrsfs }
\usepackage{graphicx,color,bm}
\usepackage{etoolbox}
\usepackage{flafter}
\usepackage{subfig}
\usepackage{tabu}
\usepackage{layout}
\usepackage{tabularx}
\usepackage{lipsum}

\let\OLDthebibliography\thebibliography
\renewcommand\thebibliography[1]{
  \OLDthebibliography{#1}
  \setlength{\parskip}{0pt}
  \setlength{\itemsep}{0pt plus 0.3ex}
}
\usepackage{floatrow}
\DeclareFloatFont{tiny}{\tiny}
\title{Heterogeneity Loss to Handle Intersubject and Intrasubject Variability in Cancer}

\iclrfinalcopy

\author{Shubham Goswami,
Suril Mehta,
Dhruva Sahrawat and
Anubha Gupta \\
SBILab, Deptt. of ECE, IIIT-Delhi, Delhi, India. \\
\texttt{\{shubham14100,suril15104,dhruva15026,anubha\}@iiitd.ac.in}
\And
Ritu Gupta \\
Laboratory Oncology Unit, Dr. B.R.A.IRCH, 
AIIMS, Delhi, India.\\
\texttt{\{drritugupta\}@gmail.com} 
}

\begin{document}
\maketitle
\vspace{-2em}
\begin{abstract}
\vspace{-1em}
Developing nations lack adequate number of hospitals with modern equipment and skilled doctors. Hence, a significant proportion of these nations' population, particularly in rural areas, is not able to avail specialized and timely healthcare facilities. In recent years, deep learning (DL) models, a class of artificial intelligence (AI) methods,  have shown impressive results in medical domain. These AI methods can provide immense support to  developing nations as affordable healthcare solutions. This work is focused on one such application of blood cancer diagnosis. However, there are some challenges to DL models in cancer research  because of the unavailability of a large data for adequate training and the difficulty of capturing heterogeneity in data at different levels ranging from acquisition characteristics, session, to subject-level (within subjects and across subjects). These challenges render DL models prone to overfitting and hence, models lack generalization on prospective subjects' data. In this work, we address these problems in the application of B-cell Acute Lymphoblastic Leukemia (B-ALL) diagnosis using deep learning. We propose heterogeneity loss that captures subject-level heterogeneity, thereby, forcing the neural network to learn subject-independent features. We also propose an unorthodox ensemble strategy that helps us in providing improved classification over models trained on 7-folds giving a weighted-$F_1$ score of 95.26\% on unseen (test) subjects' data that are, so far, the best results on the C-NMC 2019 dataset for B-ALL classification.
\end{abstract}
\vspace{-1.5em}
\section{Introduction}
\label{sec:intro}
\vspace{-1em}
Acute Lymphoblastic Leukemia (ALL) is a type of immature white blood cell cancer. As ALL progresses quickly in a few months, its early stage detection is crucial.
In populous countries such as India, even a small percentage of people diagnosed with cancer can result in a large number of people requiring urgent attention and diagnosis in early stages of cancer. Artificial intelligence (AI) can be used to build affordable and easily deployable solution to classify cancer versus normal cells. 
This work focuses on the identification of cancer and healthy cells in B-lineage ALL cancer. This cancer constitutes approx. 20\% of the pediatric cancers \cite{manoharan2009cancer}. The problem was explored earlier too \citep{baseline-1,baseline-2}, but the dataset were limited in size to less than 400 cell images. We have used C-NMC 2019 challenge dataset of IEEE ISBI 2019 that consists of nearly 14000 cell images \citep{ref50}. One of the earlier works on the C-NMC 2019 dataset was carried out by \cite{10.1007/978-3-319-66179-7_50}, who presented  stain deconvolutional layer based convolutional neural networks (CNN)  model to classify healthy and cancer cells by projecting the image data to optical density space via stain deconvolution. However, a major limitation of the approach was that the train-test split was not done at the subject-level. Hence, the images of the same subject could be present in both the training and the test data. This can cause the classifier to fail on the prospective (new unseen) subjects' data. In this paper, we conducted training and testing at the subject-level as described in the train-test splits of the C-NMC 2019 dataset, i.e., all images of a given subject will be present in either train or validation or test set. 
A number of works have been published on this dataset. \cite{isbi3} used an ensemble of CNN and RNN, \cite{isbi4} used ensemble of Inception-v2, Inception-v3, and Densenet with cross entropy loss, \cite{isbi5} used an ensemble of Densenet, Resnet and VGG, \cite{isbi7} showed the comparison of different CNN architectures and used ResNeXt (50 and 101) as classifier with GAN to augment the data, while \cite{isbi8} used Resnet with Fischer vector aggregation and neighbourhood correction. Recently, \cite{gehlot2020sdct} introduced SDCT-AuxNet$\theta$ where the classifier uses features of CNN network and those from auxiliary classifier. Also, stain deconvolved quantity images are used instead of the traditional RGB images. However, none of these works handled intersubject or intrasubject heterogeneity in the dataset, failure of which can result in  learning subject-dependent features by the classifier. In this paper, we have addressed the effect of subject-level heterogeneity.
\vspace{-1em}
\section{Dataset Description}
\label{sec:cnmc}
\vspace{-1em}
We have used C-NMC 2019 challenge dataset of IEEE ISBI 2019 that consists of segmented white blood cell images of 69 ALL cancer patients and 49 healthy subjects. This dataset is publicly available as C-NMC 2019 B-ALL classification challenge at The Cancer Imaging Archive (TCIA) \citep{ref56,ref49} with the description of distribution of subjects into train and test splits. This dataset is prepared from microscopic images captured from the bone marrow aspirate smears of subjects. The images were normalized for stain color variability  \citep{ref53} and cells, marked by expert oncologists, were segmented  \citep{ref54}.
The dataset was collected by All India Institute of Medical Sciences (AIIMS) Delhi, India. This hospital receives some of the most critical cases from across the country, with many cases referred to AIIMS by other states' regional hospitals. Hence, this dataset is extremely rich in terms of subject-level diversity that makes C-NMC 2019 dataset as one of the best available dataset for B-ALL classification.
\vspace{-1em}
\section {Motivation}
\label{sec:motivation}
\vspace{-0.75em}
This problem of the discrimination of the healthy cells from cancer cells in B-ALL cancer is very challenging owing to the following reasons: 
\vspace{-0.5em}
\begin{enumerate}
    \item Morphologically, the cells of the two classes (healthy and cancer) appear similar (Fig. 1). 
    \begin{figure}[!ht]
    \vspace{-0.5em}
    \begin{center}
      \subfloat{\includegraphics[scale=0.35, trim=0 0 -10 20]{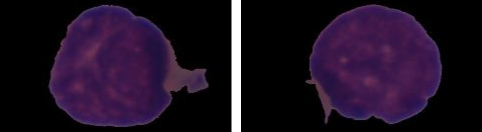}}
        \subfloat{\includegraphics[scale=0.45, trim=0 0 0 20]{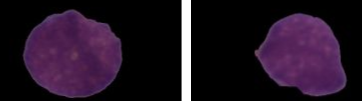}}
         \end{center}
     \label{fig:samples}
     \caption{\small LEFT: Two cancer cells, RIGHT: Two healthy cells}
     \vspace{-1.5em}
    \end{figure} 
\vspace{-0.5em}
  \item  In biomedical applications, dataset  invariably contain intersubject heterogeneity. Also, each subject's data consists of a set of images that may contain intrasubject heterogeneity. These heterogeneities can force the classifier to learn subject-level  and  subject-specific  features  instead  of  learning  class-specific  features  \citep{Allison2014}, leading to poor performance on the test set. This is to note that subject-level heterogeneity is not only limited to biomedical datasets, but is relevant wherever the dataset used for training consists of any subject-level bias, say, gender bias in datasets \citep{bias}.
Also, C-NMC 2019 dataset suffers from a long tail distribution, i.e., there is a significant difference in the number of images across subjects. This can lead to overfitting at the subject-level and can force the classifier to give more weight to subjects with more images as compared to subjects with less number of images. This can lead to poor generalization on unseen subjects.
  \vspace{-1em}
  \item There is a class imbalance in this dataset with almost double the number of images of the healthy class compared to the cancer class. 
\end{enumerate}
\vspace{-0.5em}
In order to deploy a reliable DL-based tool for B-ALL classification, we propose a deep learning based solution, addressing above challenges, in Sec-\ref{sec:pipeline} for capturing subject-level heterogeneity by: 1) data sampling strategy to capture subject-level heterogeneity in the training data;
2) new loss function, namely, heterogeneity loss for ALL cell classification that handles the fundamental problem of subject-level heterogeneity of data for better generalization of the model on unseen test data; and 3) by an unorthodox ensemble strategy that exploits confidence scores of independently trained Inception-v3 (trained on the proposed loss function) to arrive at the final decision.
\vspace{-1.25em}
\section{Proposed methodology}
\label{sec:pipeline}
\vspace{-0.75em}
\subsection{\textbf{Data Sampling}}
\vspace{-0.5em}
The data split provided in C-NMC 2019 consists of the training set of 101 subjects with a total of 12528 cell images and final test set of 17 subjects with a total of 2586 images. Test set labels are not provided and the final test scores are required to be checked via submission at the leaderboard \citep{ref50}. Due to the fewer number of subjects in the dataset, capturing of subject-level heterogeneity from the training data is hard. To address this problem, we sampled 7 different folds at the subject-level split from the training dataset in a stratified manner by maintaining the ratio of number of cancer cells to healthy cells across all the folds such that a subject's data is either present in the training set or is present in the validation set in a given fold as shown in Table-\ref{sampling}. 
Seven CNN networks are trained independently on each fold.
\begin{table}[!ht]
\caption{\small Cross-validation splits of training data into 7 folds: Number of cell images (number of subjects)} \begin{small}
\vspace{-1em}
\begin{tabular}{|l|l|l|l|l|l|l|l|l|} \hline
\textbf{Data} & \textbf{1} & \textbf{2} & \textbf{3} & \textbf{4} & \textbf{5} & \textbf{6} & \textbf{7} & \textbf{Total} \\
 \hline
Cancer & 1234 (13)& 1166 (9) & 1269 (9) & 1275 (6) & 1135 (9) & 1197 (6) & 1215 (8) & 8491 (60) \\  \hline
Healthy & 529 (9) & 546 (7) & 516 (6) & 624 (4) & 618 (6) & 603 (4) & 601 (5) & 4037 (41)\\  \hline
\end{tabular}
\end{small}
\label{sampling}
\vspace{-1em}

\end{table}
\vspace{-2em}
\subsection{\textbf{Network Architecture}}
\vspace{-0.5em}
We have used Inception-v3 deep convolution neural network (CNN) for solving this problem. 
 Different kernel sizes (from as small as $3 \times 3$ to large kernels) capture different receptive fields of the input image at a layer in CNN network \citep{peng2017large}. Inception-v3 CNN \citep{szegedy2016rethinking} with 42-layers and around 23 million parameters exploits this. It fulfils our requirement of the simultaneous use of different kernel sizes for capturing and exploiting information at different scales. We have initialized weights of Inception-v3 by pre-training on ImageNet dataset.
\vspace{-1.0em}
\subsection{\textbf{Proposed Heterogeneity Loss Function}}
 \vspace{-0.5em}
The challenges discussed in Sec-\ref{sec:motivation} have a striking resemblance to those encountered in face recognition domain. \cite{10.1007/978-3-319-46478-7_31} handled intraclass variations in face recognition by applying a loss on the learned features along with the class centers. We hypothesize that not only images belonging to separate classes should be far apart in the feature space, images belonging to the same class should also form distinct and individual compact-feature clusters. Alternatively, contrastive loss \citep{1640964,NIPS2014_5416} and triplet loss \citep{triplet_cvpr} were used on deep features to combat intraclass variations. However, these methods require image pairs or triplets, respectively, where the number of training pairs or triplets grow dramatically leading to an  increase in computational complexity. This makes the training process very inconvenient \citep{10.1007/978-3-319-46478-7_31}. On the other hand, we propose a new loss function, namely, heterogeneity loss, in equation (\ref{eq:3}) that uses multiple-instance centre loss to capture intraclass compactness along with the intersubject and intrasubject heterogeneity and does not suffer with the disadvantages of contrastive or triplet losses. 
The heterogeneity loss function is explained as follows. Let a mini-batch of size $m$ be defined as $M = \{x_i | 1 \leq i \leq m\}$, where $x_i$ represents the deep feature of the $i^\text{th}$ sample in the mini-batch $M$. Let $n_c$ and $n_s$ represent the number of classes and the number of subjects in the training set, respectively and, $S=\{i | 1 \leq i \leq n_s \}$ and $C=\{j | 1 \leq j \leq n_c \}$ denote the subject labels and class labels of the training set, respectively. We define two mappings:  a) for all ${x \in M}$, $S(x)$ represents the subject label of $x$, and b) for all ${x \in M \lor x\in S}, C(x) \in C $ represents the class label over deep feature $x$ and subject label $s \in S$.
We define mini-batch subsets of $M$ on class-level and subject-level as follows: for all ${c \in C}$,  $M_c = \{x | x\in M \land C(x)=c\}$ s.t. $\cup_{c \in C} M_c=M$ and for all ${s \in S}$, $M_s = \{x | x\in M \land S(x)=s\}$ s.t. $\cup_{s \in S} M_s=M$. $c_{1,c}$  represents the class centre for $c \in C$ and $c_{2,s}$ represents the subject centre for $s \in S$. Next, we define the heterogeneity loss function  $\mathcal{L}_{H}$ as:
\begin{equation}\label{eq:3}
\vspace{-2mm}
   \mathcal{L}_{H} = \mathcal{L}_{CE} +
   \lambda_{1}\mathcal{L}_{C_{class}}+
   \lambda_{2}\mathcal{L}_{C_{subject}}+ \lambda_{3}\mathcal{L}_{C_{subject-class}}, \text{  where}
\end{equation}   
\begin{equation}
\label{eq:2}
\mathcal{L}_{CE} = -\frac{1}{|M|}\sum_{x \in M}\log\dfrac{e^{W_{C(x)}^{T}x+b_{C(x)}}}
{\sum_{c\in C}e^{W_{c}^{T}x+b_{c}}}, 
\end{equation}
\begin{equation}
\label{eq:class_c}
\mathcal{L}_{C_{class}}=\sum_{c \in C} \dfrac{1}{|M_c|}\sum_{x \in M_c}\norm{x - c_{1,c}}_{2}^{2},
\end{equation}
   \begin{equation}\label{eq:4}
   \mathcal{L}_{C_{subject}}=\sum_{\substack{s \in S \\ s.t. \\ |M_s| > 0}} \dfrac{1}{|M_s|}\sum_{x \in M_s}\norm{x - c_{2,s}}_{2}^{2},
\end{equation}  
\begin{equation}\label{eq:6}
\mathcal{L}_{C_{subject-class}}= \sum_{\substack{s_i \in S \\ s.t. \\ |M_{s_i}|\neq 0}} \hspace{5pt} \sum_{\substack{s_j \in S \\ s.t. \\ |M_{s_j}|\neq0 \\ \land s_i\neq s_j}}\begin{cases}
    \norm{c_{2,s_i} - c_{2,s_j}}_{2}^{2},& \text{if }{C(s_i)= C(s_j)} \vspace{8pt}\\
    
    \dfrac{1}{1+\norm{c_{2,s_i} - c_{2,s_j}}_{2}^{2}},  & \text{otherwise}\
\end{cases}
\end{equation}
where $\lambda_1$, $\lambda_2$ and $\lambda_3$ are the hyper-parameters associated with $\mathcal{L}_{H}$.
Centre loss was introduced by  \cite{10.1007/978-3-319-46478-7_31}  to increase interclass distance and intraclass compactness to learn better discriminative features of the output class labels. $\mathcal{L}_{H}$ consists of two components.
The first component handles output class inference and consists of two losses: softmax-cross entropy loss $\mathcal{L}_{CE}$ shown in  (\ref{eq:2}) and class centre loss $\mathcal{L}_{C_{class}}$ shown in  (\ref{eq:class_c}) that handles the inference. The second component tries to capture the heterogeneity at the subject-level for learning subject independent features by using centre loss as a multi-task auxiliary loss. This is achieved by the weighted sum of  $\mathcal{L}_{C_{subject}}$ shown in  (\ref{eq:4}) and $\mathcal{L}_{C_{subject-class}}$ shown in  (\ref{eq:6}). This second component handles intersubject and intrasubject heterogeneity by geometrically forcing the output of the last fully connected layer because $\mathcal{L}_{C_{subject}}$ increases intrasubject compactness for each subject and hence, learns features common to all images belonging to the same subject. $\mathcal{L}_{C_{subject-class}}$ decreases intersubject distance between the subjects belonging to the same class.
Increase in intra-subject compactness and decrease in  inter-subject distance at the same time is important because it forces the network to learn subject-independent features.  $\mathcal{L}_{C_{class}}$  defined in  (\ref{eq:class_c}) is the sum of mean centre loss with respect to each output class label $c$ over $M$. The mean term $\dfrac{1}{|M_c|}$ for $c \in C$ serves as class weights to handle class imbalance. Similarly, $\mathcal{L}_{C_{subject}}$ can handle subject imbalance for a given mini-batch. 
\vspace{-1em}
\section{Experiments and Results}
\vspace{-0.5em}
\subsection{Training}
\begin{figure*}[!ht]
\vspace{-2.0em}
\begin{center}
\subfloat{\includegraphics[width=0.235\textwidth]{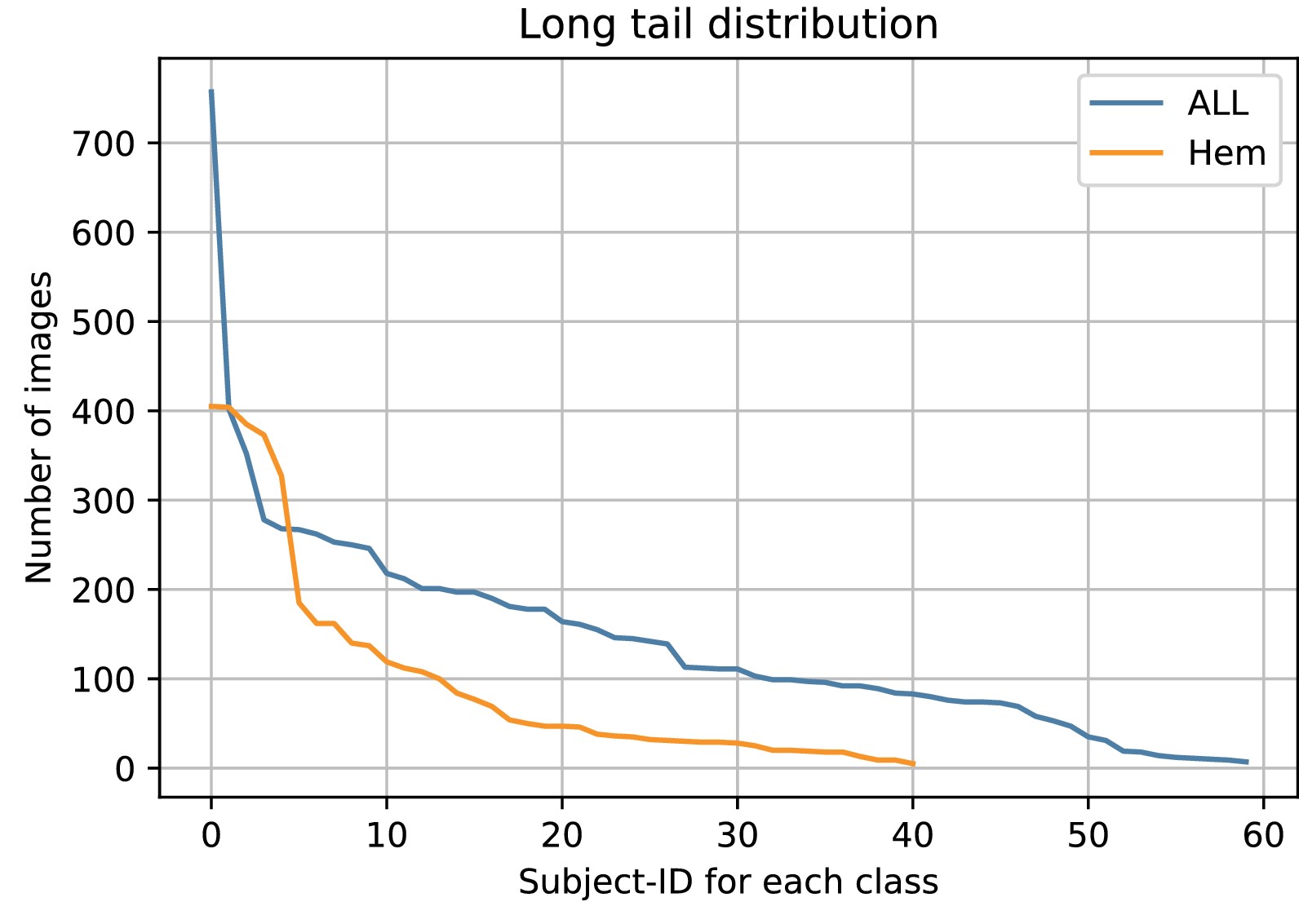}}
\subfloat{\includegraphics[width=0.265\textwidth]{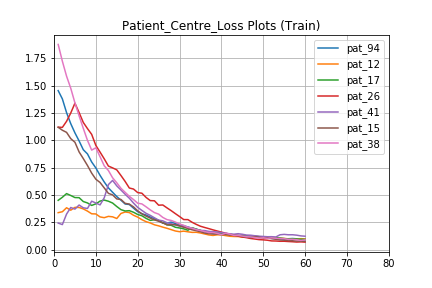}}
\subfloat{\includegraphics[width=0.265\textwidth]{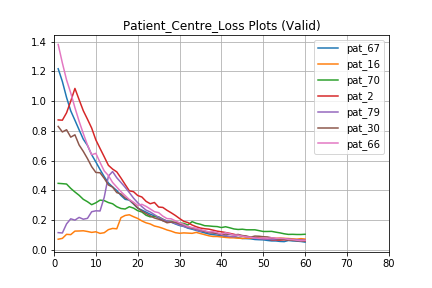}}
\subfloat{\includegraphics[width=0.215\textwidth]{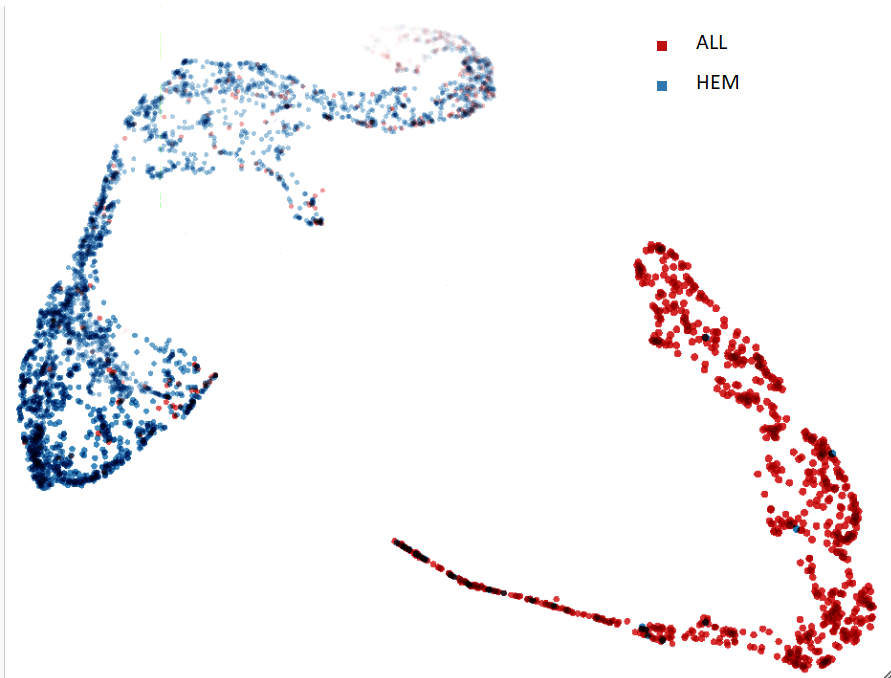}} 
\end{center}
\caption{\small Left to Right: Long tail distribution of dataset, training, validation convergence plots for \protect$\mathcal{L}_{C_{subject}}$ for random subjects sampled for training and validation set and T-SNE visualization of class-level on validation set of a randomly selected fold out seven folds. Convergence plots (with smoothness=0.8). 
}
\label{fig:det}
\vspace{-1em}
\end{figure*}

\vspace{-0.5em}
We used Pytorch along with NVIDIA GTX 1080 Ti for training. Single Inception-v3 model trains in around 4 hours. We used both Train time Augmentation (TrA) and Test time Augmentation (TTA).  Initially, all the images were cropped to a fixed size of $400 \times 400$ pixels. For robust training, several augmentation techniques were applied such as HorizontalFlip, VerticalFlip, RandomRotation, Affine Transformation, adjustment of hue, saturation, contrast, brightness, and RandomCrop that resizes the images to 
$299 \times 299$ for Inception-v3. 
For each fold we used \textbf{two stage training} of Inception-v3 model on $\mathcal{L}_{H}$ loss with $\lambda_1$, $\lambda_2$ and $\lambda_3$ with values $0.05$, $0.05$ and $0.005$, respectively (tuned over multiple experiments), Adam optimizer and a batch size of 16. The first stage used a learning rate of $lr=10^{-3}$ and the best model was saved before the loss started diverging on the training set. In the second stage, we resumed training from the first phase's checkpoint with $lr=10^{-5}$. 
We also used \textbf{Test time augmentation (TTA)}, called as 5-crop strategy with $229 \times 299$ size crops: one each from the four corners and one from the centre of the unseen input cell image. 
Finally, average confidence scores on these 5 crops for each input image is calculated. Comparative performance of different loss configurations in Table-\ref{table:loss} shows that heterogeneity loss indeed performs best. Fig-\ref{fig:detA} shows the accuracy and convergence plots of $\mathcal{L}_{H}$ loss on independently trained Inception-v3 model corresponding to each of the 7 folds. Fig-\ref{fig:det} shows convergence of $\mathcal{L}_{C_{subject}}$ on subjects randomly sampled from the training and validation sets.  Convergence of $\mathcal{L}_{C_{subject}}$ is important because $\mathcal{L}_{C_{subject-class}}$ is formulated using subject centers and also it shows that the classifier can handle class-subject imbalance. Overall convergence of $\mathcal{L}_{H}$ would lead to a good generalization of the model on the unseen test data. Fig-\ref{fig:det} also shows class-level T-SNE visualization (on the same random split as above) using the output of last layer of Inception-v3 as features.

\begin{table}[!ht]
\caption{\small Performance of Inception-v3 with different loss configurations using training and validation sets provided on the challenge portal \citep{ref50}}

\vspace{-1em}
\begin{center}
\begin{small}
\begin{tabular}{|c|c|l|}
\hline
\textbf{Loss Function}                         		 & \textbf{Accuracy} (\%) & \textbf{Weighted}-$F_1$ (\%) \\
\hline
$\mathcal{L}_{CE}$                                          		& 84.66         & 84.25         \\ \hline
$\mathcal{L}_{CE}+\mathcal{L}_{C_{class}}$                              	& 86.14           & 85.84         \\ \hline
$\mathcal{L}_{CE}+\mathcal{L}_{C_{subject}} +\mathcal{L}_{C_{class}}$ 	& 88.88           & 88.62           \\ \hline
$\mathcal{L}_{H}=\mathcal{L}_{CE} +\mathcal{L}_{C_{subject}}$ $+\mathcal{L}_{C_{class}} +  \mathcal{L}_{C_{subject-class}}$
&	 \textbf{90.05}           & \textbf{89.94}      \\  
\hline
\end{tabular}
\end{small}
\end{center}

\label{table:loss}
\vspace{-1.5em}
\end{table}
\vspace{-1em}
\subsection{Ensemble-based Decision}
\vspace{-0.5em}
One Inception-v3 model was trained on each of the  7-folds. Table-\ref{table:folds} shows that all of these 7 models provide good accuracy and weighted-$F_1$ score on their respective validation sets. Since these seven Inception-v3 models are not weak classifiers, one can use max-voting or average-voting on the outputs of these classifiers to make the final decision. To achieve this, we propose an intuitive ensembling that uses both max-voting and average-voting in a novel way.  First, we consider the maximum confidence score on a test sample from the output of all seven classifiers. If this confidence score is more than a certain threshold $\theta(\geq 0.90)$, we consider it as the final decision and hence, use max-voting. If this confidence score is below $\theta$, we consider the harmonic mean of all the seven confidence scores as the final decision. 
We experimented with two values of $\theta$ above 0.9 ($\theta \in \{0.95,0.98\}$, explained in detail in Sec-\ref{sec:theta}).
This is to note that the detection of true positives and true negatives is equally important in cancer diagnosis because a subject having cancer if left untreated may die. Likewise, a subject not having cancer, if stated to be suffering with cancer, may have traumatic and financial impact on the subject and family, besides health injury for giving unnecessary chemotherapy. Weighted-$\text{F}_1$ score is a reliable metric to evaluate the overall classifier’s performance, especially, when there is class imbalance in the dataset as it calculates $F_1$ score for each output class label and finds their average weighted by support (the number of true instances for each label). This alters ‘macro’ and 'micro' $F_1$ score to account for label imbalance, which is also one of the challenges associated with this dataset. 
\vspace{-1.0em}
\section{Concluding Remarks}
\vspace{-0.5em}
As shown in Table-\ref{table:final}, we achieved a weighted $F_1$ score of 95.26\% on the final test. (Refer to the submission on the leaderboard \citep{ref50} with the username: \textit{shubham14100}). Various recent research works discussed in Sec-1 used superior CNN architectures, say Resnet, ResNext, Densenet \citep{isbi7} and ensemble of these architectures \citep{isbi4} using cross-entropy loss to evaluate weighted $F_1$ score on CNMC-2019 dataset. However, our weighted $F_1$ score is so far highest on this dataset as reported in Table-\ref{table:final} because we have addressed the fundamental issue of heterogeneity in data by handling intersubject and intrasubject variability (subject-level heterogeneity) via incorporating it in sampling and the heterogeneity loss  function ($\mathcal{L}_{H}$). Further, the ensemble approach gives better weighted $F_1$ score because this technique makes the overall inference more reliable as compared to the individual inference as shown in Table-\ref{table:folds}.
Our results indicate that the proposed method successfully handles intrasubject and intersubject heterogeneity and generalizes better on unseen data. Hence, the proposed method can be extended in a similar fashion to other applications that encounter some type of subject level heterogeneity in the dataset.
\begin{table}[!ht]
\caption{\small Overall comparison on CNMC-2019 final test set.}

\begin{center}
\vspace{-1em}
\begin{small}
\begin{tabular}{|l|c|c|c|b|} \hline
 \multicolumn{2}{|c|}{\textbf{Comparison of different works}}  & \multicolumn{1}{c|}{\textbf{Weighted-$F_1$(\%)}}  \\ \hline
 \multicolumn{2}{|c|}{\cite{isbi7}} & 84.90 \\ \hline
 \multicolumn{2}{|c|}{\cite{isbi4}} & 85.52 \\ \hline
\multicolumn{2}{|c|}{\cite{isbi3}} & 86.6 \\ \hline
\multicolumn{2}{|c|}{\cite{isbi5}} & 87.98 \\ \hline
\multicolumn{2}{|c|}{\cite{isbi8}} & 91.04 \\ \hline
\multicolumn{2}{|c|}{\cite{gehlot2020sdct}} & 94.86 \\ \hline
\textbf{Proposed} & \multicolumn{1}{c|}{$\theta = 0.95$} & \textbf{95.26}\\ \cline{2-3} 
(ensemble with harmonic mean) &  \multicolumn{1}{c|}{$\theta = 0.98$} & \textbf{95.24} \\ \hline
\end{tabular}
\end{small}
\end{center}
\label{table:final}

\vspace{-2em}
\end{table}

\vspace{-0em}
\subsubsection*{Author Contributions}
\vspace{-0.5em}
S.G., S.M. and D.S. contributed equally. A.G. and R.G. arranged funding and prepared the dataset. S.G., S.M., and D.S. proposed the method, coded and generated results. All authors discussed the results and prepared the manuscript.

\vspace{-0em}
\subsubsection*{Acknowledgments}
\vspace{-0.5em}
Authors  gratefully  acknowledge  the  research  funding  support (Grant: EMR/2016/006183) from the Department of Science and Technology, Govt.of India for this research work. Authors would also like to acknowledge the support from Infosys Centre for AI, IIIT-Delhi, India.

\bibliography{iclr2020_conference}
\bibliographystyle{iclr2020_conference}
\newpage
\appendix
\section{Appendix}
\label{sec:app}
\setcounter{table}{0}
\renewcommand{\thetable}{A\arabic{table}}
\setcounter{figure}{0}
\renewcommand{\thefigure}{A\arabic{figure}}

\begin{figure*}[!ht]
\vspace{-2.0em}
\begin{center}
\subfloat{\includegraphics[width=0.245\textwidth]{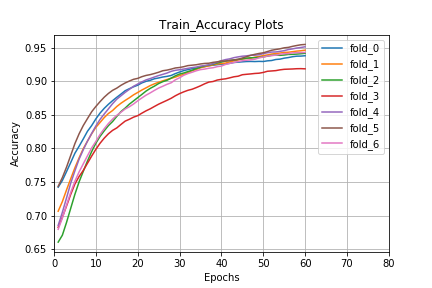}}
\subfloat{\includegraphics[width=0.245\textwidth]{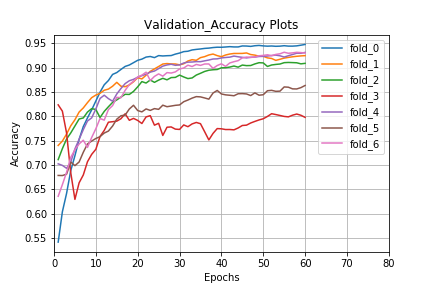}}
\subfloat{\includegraphics[width=0.245\textwidth]{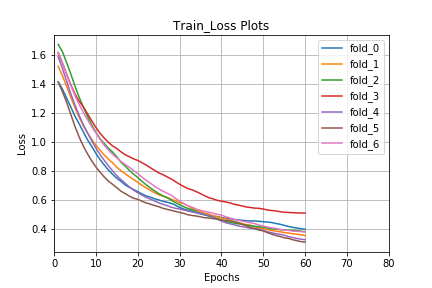}}
\subfloat{\includegraphics[width=0.245\textwidth]{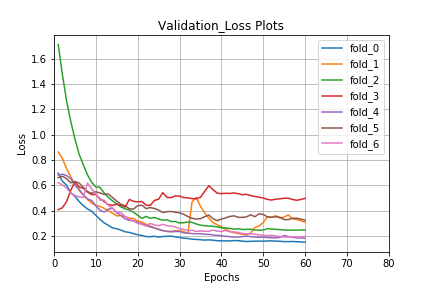}} \\
\vspace{-1.0em}
\subfloat{\includegraphics[width=0.245\textwidth]{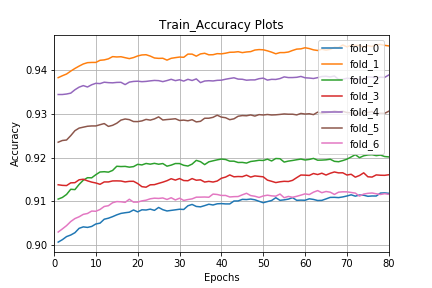}}
\subfloat{\includegraphics[width=0.245\textwidth]{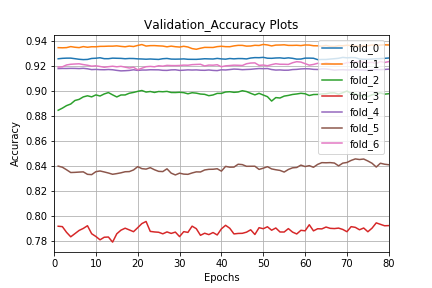}}
\subfloat{\includegraphics[width=0.245\textwidth]{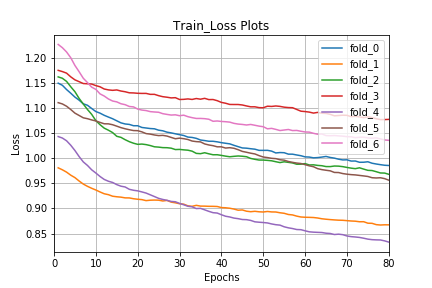}}
\subfloat{\includegraphics[width=0.245\textwidth]{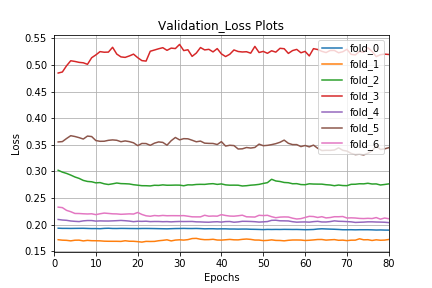}}
\vspace{-0.25em}
\end{center}
\caption{
\textbf{TOP}: Left to Right: Phase-I training with lr=$1e-3$, \textbf{BOTTOM}: Left to Right: Phase-II training with lr=$1e-5$ . For both rows from Left to Right: training accuracy, validation accuracy, training loss and validation loss respectively, where loss is $\mathcal{L}_{H}.$ 
}
\label{fig:detA}
\vspace{-1em}
\end{figure*}

\begin{table}[!ht]
\caption{ Overall comparison of each fold's model with the final proposed ensemble output }

\begin{center}
\vspace{-1em}
\begin{small}
\begin{tabular}{|l|c|c|c|c|b|} \hline
\textbf{} & \multicolumn{2}{c|}{\textbf{on validation}} & \multicolumn{1}{c|}{\textbf{on test (with TTA)}}  \\ \hline
model & \textbf{Accuracy (\%)} & \textbf{Weighted-$F_1$(\%)}  & \textbf{Weighted-$F_1$(\%)} \\ \hline
Fold-0 model & 95.25 & 95.26 &   92.72 \\ \hline
Fold-1 model & 94.60 & 94.59 &   94.41 \\ \hline
Fold-2 model & 92.97 & 92.97 &   93.63 \\ \hline
Fold-3 model & 83.78 & 89.58 &   92.13 \\ \hline
Fold-4 model & 94.96 & 94.96 &   91.43 \\ \hline
Fold-5 model & 86.81 & 86.82 &   89.82 \\ \hline
Fold-6 model & 94.492 & 94.93 &   90.39 \\ \hline
\textbf{Proposed} & \multicolumn{2}{c|}{$\theta = 0.95$} & \textbf{95.26}\\ \cline{2-4} 
(ensemble with harmonic mean) &  \multicolumn{2}{c|}{$\theta = 0.98$} & \textbf{95.24} \\ \hline
\end{tabular}
\end{small}
\end{center}
\label{table:folds}

\vspace{-2em}
\end{table}

\subsection{Choosing $\theta$}
\label{sec:theta}
The threshold $\theta$ is chosen as follows. First, we arbitrarily chose two high confidence values above 0.9 as 0.95 and 0.98. For each $\theta \in \{0.95.0.98\}$ in Table-\ref{a1}, we report the accuracy and weighted-$F_1$ on a subset of validation set corresponding to each fold in which the confidence output by their respective models is greater than  $\theta$. It is clear from Table-\ref{a1} that we can confidently use max-voting if the maximum confidence score for a test example is more than $\theta$. We used harmonic mean for averaging the confidence scores if none of them is more than $\theta$ as harmonic mean is more stable towards outliers.
\begin{table}[!ht]
\caption{\small Accuracy and Weighted-$F_1$ score on validation data having output confidence score more than $\theta$, for each fold and $\theta \in \{0.95,0.98\}$}
\begin{small}
\begin{tabular}{|l|c|c|c|} \hline
 & &  \textbf{Accuracy} (\%) & \textbf{Weighted-$F_1$} (\%) \\ \hline
Fold-0 model & $\theta=0.95$ & 98.74 & 98.75 \\ \cline{2-4}
& $\theta=0.98$ & 99.26 & 99.27 \\ \hline
Fold-1 model & $\theta=0.95$ & 99.08 & 99.08 \\ \cline{2-4}
& $\theta=0.98$ & 99.39 & 99.39 \\ \hline
Fold-2 model & $\theta=0.95$ & 99.25 & 99.24 \\ \cline{2-4}
& $\theta=0.98$ & 99.61 & 99.62 \\ \hline
Fold-3 model & $\theta=0.95$ & 99.22 & 99.23 \\ \cline{2-4}
& $\theta=0.98$ & 99.65 & 99.65 \\ \hline
Fold-4 model & $\theta=0.95$ & 99.25 & 99.26 \\ \cline{2-4}
& $\theta=0.98$ & 98.11 & 98.11 \\ \hline
Fold-5 model & $\theta=0.95$ & 97.20 & 97.30 \\ \cline{2-4}
& $\theta=0.98$ & 98.01 & 98.02 \\ \hline
Fold-6 model & $\theta=0.95$ & 98.53 & 98.55 \\ \cline{2-4}
& $\theta=0.98$ & 98.07 & 97.12 \\ \hline
\end{tabular}
\end{small}
\label{a1}

\end{table}

\end{document}

%% file: math_commands.tex
\usepackage{amsmath,amsfonts,bm}

\def\eqref#1{equation~\ref{#1}}

\def\1{\bm{1}}

\DeclareMathAlphabet{\mathsfit}{\encodingdefault}{\sfdefault}{m}{sl}
\SetMathAlphabet{\mathsfit}{bold}{\encodingdefault}{\sfdefault}{bx}{n}

